\documentclass[10pt,twocolumn,letterpaper]{article}

\usepackage[pagenumbers]{cvpr} 

\usepackage{algorithm}
\usepackage{amsmath}
\usepackage{amssymb}
\usepackage{amsthm}
\usepackage{booktabs}
\usepackage{color}
\usepackage{enumitem}
\usepackage{graphicx}
\usepackage{listings}
\usepackage{makecell}
\usepackage{mathrsfs}
\usepackage[free-standing-units]{siunitx}
\usepackage{subcaption}
\usepackage{url}
\usepackage{wrapfig}
\usepackage{xcolor}

\usepackage{bbding}

\definecolor{deepblue}{rgb}{0,0,0.5}
\definecolor{deepred}{rgb}{0.6,0,0}
\definecolor{deepgreen}{rgb}{0,0.5,0}

\definecolor{es-blue}{rgb}{0,0.4,0.8}

\usepackage{color}
\usepackage{xcolor}
\definecolor{citecolor}{HTML}{0071bc}
\usepackage[pagebackref=true,breaklinks=true,colorlinks,citecolor=citecolor,bookmarks=false]{hyperref}

\usepackage[capitalize]{cleveref}
\crefname{section}{Sec.}{Secs.}
\Crefname{section}{Section}{Sections}
\Crefname{table}{Table}{Tables}
\crefname{table}{Tab.}{Tabs.}

\def\cvprPaperID{****} 
\def\confName{CVPR}
\def\confYear{2023}

\begin{document}

\title{Neural Volumetric Memory for Visual Locomotion Control}

\author{
Ruihan Yang\textsuperscript{1} \hspace{2mm}
Ge Yang\textsuperscript{2,3} \hspace{2mm}
Xiaolong Wang\textsuperscript{1} \hspace{2mm} \\
\vspace{1mm}
\textsuperscript{1}UC San Diego\;
\textsuperscript{2}Institute of AI and Fundamental Interactions\;
\textsuperscript{3}MIT CSAIL
}

\twocolumn[{%
\vspace{-1em}
\maketitle
\vspace{-1em}
\begin{center}
    \centering 
    \vspace{-0.3in}
    \includegraphics[width=\textwidth,clip]{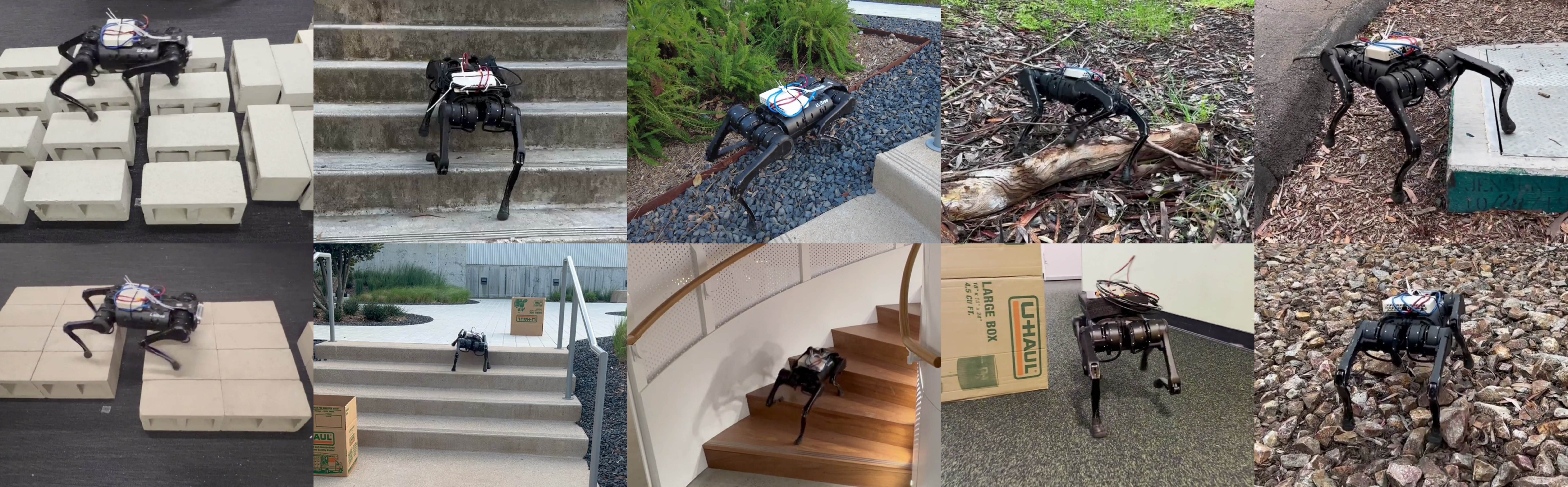}
    \vspace{-0.15in}
    \captionof{figure}{\textbf{Real-World Deployment:} We deploy our policy in real-world environments with Stones, Stairs, Stages, Obstacles, and Unstructured Terrain. Our robot utilizes the volumetric memory of the surrounding 3D structure to successfully traverse complex environments.
    }\label{fig:teaser}
    \vspace{10pt}
\end{center}
}]

\begin{abstract}
Legged robots have the potential to expand the reach of autonomy beyond paved roads. In this work, we consider the difficult problem of locomotion on challenging terrains using a single forward-facing depth camera. Due to the partial observability of the problem, the robot has to rely on past observations to infer the terrain currently beneath it. To solve this problem, we follow the paradigm in computer vision that explicitly models the 3D geometry of the scene and propose Neural Volumetric Memory (NVM), a geometric memory architecture that explicitly accounts for the SE(3) equivariance of the 3D world. NVM aggregates feature volumes from multiple camera views by first bringing them back to the ego-centric frame of the robot. We test the learned visual-locomotion policy on a physical robot and show that our approach, which explicitly introduces geometric priors during training, offers superior performance than more na\"ive methods. We also include ablation studies and show that the representations stored in the neural volumetric memory capture sufficient geometric information to reconstruct the scene. Our project page with
videos is \url{https://rchalyang.github.io/NVM}
\end{abstract}
\vspace{-0.2in}

\section{Introduction}
\label{sec:intro}

\begin{figure*}[th]
\centering
    \includegraphics[width=\textwidth,clip]{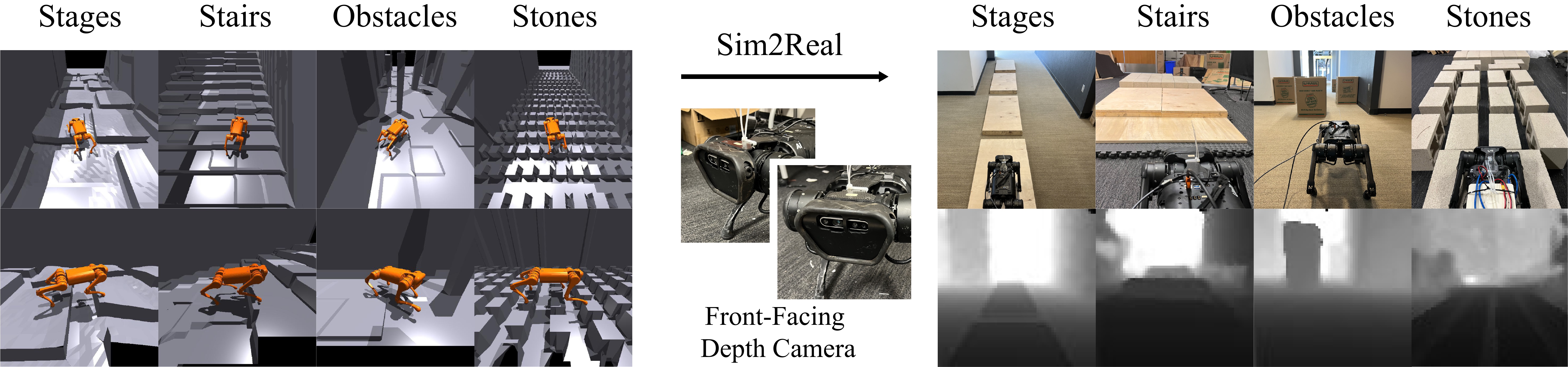}
    \vspace{-0.15in}
    \caption{\small{\textbf{Overview of Simulated Environment \& Real World Environment:} Our simulated environments are shown on the left and real-world environments are shown on the right. For the real-world environment, the corresponding visual observations for each real-world environment are shown in the bottom row. All policies are trained in the simulation and transferred into the real world without fine-tuning.
    }}\label{fig:sim2real_pipeline}
\vspace{-0.15in}
\end{figure*}

\noindent Consider difficult locomotion tasks such as walking up and down a flight of stairs and stepping over gaps (Figure~\ref{fig:teaser}). The control of such behaviors requires tight coupling with perception because vision is needed to provide details of the terrain right beneath the robot and the 3D scene immediately around it. This problem is also partially-observable. Immediately relevant terrain information is often occluded from the robot's current frame of observation, forcing it to rely on past observations for control decisions. For this reason, while blind controllers that are learned in simulation using reinforcement learning have achieved impressive results in agility and robustness~\cite{lee2020learning,Kumar2021,margolis2022rapid}, there are clear limitations on how much they can do. How to incorporate perception into the pipeline to produce an integrated visuomotor controller thus remains an open problem.

A recent line of work combines perception with locomotion using ego-centric cameras mounted on the robot. 
The predominant approach for addressing partial observability is to do frame-stacking, where the robot maintains a visual buffer of recent images. This na\"ive heuristic suffers from two major problems: first, frame-stacking on a moving robot ignores the equivariance structure of the 3D environment, making learning a lot more difficult as policy success now relies on being able to learn to account for spurious changes in camera poses. A second but subtler issue is that biological systems do not have the ability to save detailed visual observations pixel-by-pixel. These concerns motivate the creation of an intermediary, short-term memory mechanism 
to functionally aggregate streams of observation into a single, coherent representation of the world.

Motivated by these observations, we introduce a novel volumetric memory architecture for legged locomotion control. Our architecture consists of a 2D to 3D feature volume encoder, and a pose estimator that can estimate the relative camera poses given two input images. When combined, the two networks constitute a Neural Volumetric Memory (NVM) that takes as input a sequence of depth images taken by a forward-looking, ego-centric camera and fuses them together into a coherent latent representation for locomotion control. We encourage the memory space to be $SE(3)$ equivariant to changes in the camera pose of the robot by incorporating translation and rotation operations based on estimated relative poses from the pose network. This inverse transformation allows NVM to align feature volumes from the past to the present ego-centric frame, making both integrating over multiple timesteps into a coherent scene representation and learning a policy, less difficult.

Our training pipeline follows a two-step teacher-student process where the primary goal of the first stage is to produce behaviors in the form of a policy. After training completes, this policy can traverse these difficult terrains robustly, but it relies on privileged sensory information such as an elevation map and ground-truth velocity. Elevation maps obtained in the real world are often biased, incomplete, and full of errors~\cite{Miki2022-to}, whereas ground-truth velocity information is typically only available in instrumented environments. Hence in the visuomotor distillation stage of the pipeline, which still runs in the simulator, we feed the stream of ego-centric views from the forward depth camera into the neural volumetric memory. We feed the content of this memory into a small policy network and train everything end-to-end including the two network components of the NVM using a behavior cloning loss where the state-only policy acts as the teacher. For completeness, we offer an additional self-supervised learning objective (Figure~\ref{fig:SSL}) that relies on novel-view consistency for learning. The end product of this visuomotor distillation pipeline is a memory-equipped visuomotor policy that can operate directly on the UniTree A1 robot hardware (see~Figure~\ref{fig:teaser}). A single policy is used to handle all environments covered by this paper. We provide comprehensive experiment and ablation studies in both simulation and the real world, and show that our method outperforms all baselines by a large margin. It is thus essential to model the 3D structure of the environment. 

\section{Related Work}\label{sec:related_work}

\begin{figure*}[]
    \includegraphics[width=\textwidth,clip]{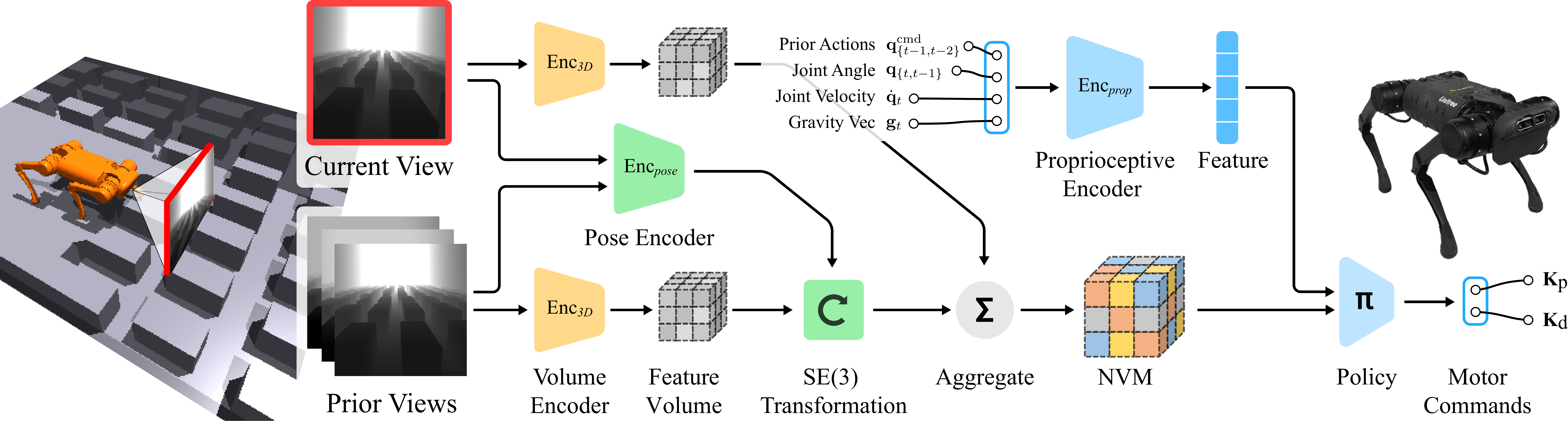}
    \vspace{-0.1in}
    \caption{\small{\textbf{N}eural \textbf{V}olumetric \textbf{M}emory (\emph{NVM}): Our NVM module extracts 3D feature volumes from every visual observation in the near past, transform them with corresponding estimated transformation into present robot frame, and fuse the transformed feature volumes into a neural volumetric memory for decision-making.}}
    \label{fig:NVM}
\vspace{-0.1in}
\end{figure*}

\paragraph{Blind Locomotion Controllers:} Legged locomotion using the proprioception of the robot has been studied for decades ~\cite{geyer2003positive,yin2007simbicon, TvdP-gi98,miura1984dynamic,bledt2018cheetah,raibert1984hopping}. Traditionally, controllers are developed with model-based methods~\cite{mitcheetah2018mpc,gaitcontroller2013,di2018dynamic,ding2019real,bledt2020extracting, grandia2019frequency,sun2021online,trajectoryopt2019}. However, most model-based methods fail to generalize to unseen environments in the real world. In order to attain better generalization and robustness, recent works~\cite{RLloco2004,tan2018sim2real,hwangbo2019learning,lee2020learning,pmtg,jain2019hrl,glide2021,Kumar2021, RoboImitationPeng20} utilize model-free reinforcement learning to train neural network controllers in simulation and directly transfer the learned policies to the real robot. These methods are able to achieve high levels of agility and adapt over diverse terrains using just the proprioceptive state of the robot. The real-world utility of these robots, however, is also limited, since these blind controllers are unable to perceive the surrounding. In this paper, we propose to integrate perception to facilitate control, with a focus on a better 3D understanding of the environments. 

\paragraph{Locomotion with Exteroceptive Perception:} The elevation map~\cite{fankhauser2018probabilistic, kweon1992high,pan2019gpu, Fankhauser2014RobotCentricElevationMapping} around the robot has been used to perform optimization and planning for foot placement~\cite{RobustRough2018, zucker2011optimization,wermelinger2016navigation, chilian2009stereo,xie2020allsteps, peng2017deeploco}. Recent RL controller~\cite{Miki2022-to, rudin2022learning} also uses elevation maps in an end-to-end manner. Using the elevation map to represent the surrounding is straightforward for foot placement optimization, but the elevation map obtained in the real world is either noisy or requires a complicated hardware pipeline. Another option is using depth or RGB images to provide visual observation for the controller. Hierarchical control systems~\cite{jain2020pixels,margolis2021learning, seo2022learning,sorokinlearntonavi2022,visual-loco-complex} are proposed to utilize depth reading while the low-level controllers still rely on only the proprioceptive state of the robot. To get the most out of the depth image, Yang et al.~\cite{yang2022learning} and Imai et al.~\cite{Imai2021VisionGuidedQL} propose to learn visual RL policy to output the raw joint angles and enable the robot to maneuver in the wild avoiding different kinds of obstacles. Agarwal et al.~\cite{agarwal2022legged} and Loquercioo et al.~\cite{loquercioccross-modal2022} train a policy using RL to traverse a large variety of terrains.  Though these works show encouraging results in specific tasks, they have not fully exploited the 3D structure of the environment. On the other hand, we propose to train a novel neural volumetric memory together with policy learning, which allows our robot to walk through more challenging terrains. 

\paragraph{Volumetric Cognitive Map: } How to represent and understand the 3D scene has been studied by the computer vision community for a long time. Hartley et al.~\cite{hartley2003multiple} and Wiles et al.~\cite{wiles2020synsin} provide representation built on geometry principles and camera motions. Learning-based approaches are proposed ~\cite{jimenez2016unsupervised,zhou2017unsupervised,tulsiani2017multi,kanazawa2018learning,wiles2020synsin,hu2021worldsheet,Rockwell2021, Lai21a} to learn 3D representation from 2D inputs. Neural 3D representations ~\cite{Tung_2019_CVPR,harley_viewcontrast,sitzmann2019deepvoxels,nguyen2019hologan,mustikovela2020self, mildenhall2020nerf, sitzmann2019scene} has also shown great result in multiple computer vision tasks ~\cite{Tung_2019_CVPR,harley2020tracking,harley_viewcontrast}. Our work is closely related to Lai et al.~\cite{Lai21a} where the deep voxel feature representing the entire scene and the camera trajectory are learned from video clips without odometry in a self-supervised manner. Instead of focusing on view synthesis, we aggregate the 3D feature volume to form a short-term memory representation for decision-making. 

\paragraph{Teacher-Student Training:} 
To accelerate training and utilize the privileged information in simulation, Miki et al. ~\cite{Miki2022-to} follow a teacher-student training framework, where the teacher policy is trained with privileged information and then distilled  into student policy where the privileged information is not available. Multiple works in different research domains ~\cite{chen2020learning, janai2020computer, kaufmann2018deep, kaufmann2020deep, chen2022system, mu2021maniskill, pinto2017asymmetric} follow this framework as well. We share a similar pipeline as ~\cite{agarwal2022legged, margolis2021learning} where elevation-map-based policies are trained first and the elevation-map-based policies are distilled visual policies afterward.

\section{Volumetric Memory for Legged Locomotion}

\noindent Legged locomotion using ego-centric camera views is intrinsically a partially-observed problem. To make the control problem tangible, our robot needs to aggregate information from previous frames and correctly infer the occluded terrain underneath it. During locomotion, the camera mounted directly on the robot chassis undergo large and spurious changes in pose, making integrating individual frames to a coherent representation non-trivial.

To account for these camera pose changes, we propose neural volumetric memory (NVM) --- a 3D representation format for scene features. It takes as input a sequence of visual observations and outputs a single 3D feature volume representing the surrounding 3D structure.

\subsection{Baking-in $\boldsymbol{SE(3)}$ Equivariance in NVM}

\noindent Our training procedure encourages the learned memory space to be $SE(3)$ equivariant by applying translation and rotation transformations on the feature volume before aggregating additional information from new observations. 

As the first step (Figure~\ref{fig:NVM}), an encoder network receives a sequence of consecutive observations 
$O_{t} .. O_{t-n+1}$. Instead of generating a compact feature embedding in the shape of a tuple, the 3D encoder $Enc_{3D}$ extracts \textbf{feature volumes} ${V}_{t} .. {V}_{t-n+1}$. A second network $Enc_{\textrm{pose}}$ estimates the relative changes in camera pose between the present frame ${O}^{v}_{t}$ and frames ${O}^{v}_{t} .. {O}^{v}_{t-n+1}$ from the past. This sequence of estimate allows NVM to transform the feature volume ${V}_{t} .. {V}_{t-n+1}$ from the past to the present frame using the transformations ${T}^{t}_{t}..{T}^{t}_{t-n+1}$. We annotate the transformed feature volume as $\hat{V}_{t} .. \hat{V}_{t-n+1}$. These canonicalized volumes can now be aggregated together by element-wise sum to produce a fused feature volume $\hat{V}^{m}_{t}$ for the surrounding 3D structure. 
This scheme allows NVM to disentangle the symmetry structure of the ambient 3D space from the ``content'' specific to the scene. By doing so, we encourage \(SE(3)\) equivariance in the memory space. 

We provide a detailed description of all components in our NVM module in the remaining section.

\noindent\textbf{3D Encoder:} Our 3D encoder $Enc_{3D}$ takes a 2D depth image as input and outputs a 3D feature volume. We parameterize this encoder with a set of 2D convolution layers followed by 3D convolution layers. 
Specifically, we generate a 2D feature map with the shape $(C, H, W)$ using 2D convolution, then reshape the feature map channel-wise into a volume tensor with the shape $(C / D, D, H, W)$. We chose a voxel-grid size of $D=12, H=6, W=6$. The 3D convolution layers are added, to further refine the 3D features.
\begin{equation*}
    V_{i} = Enc_{3D}(O^{v}_{i}) \ \ \ \ \  i=t .. t-n+1
\end{equation*}

\noindent\textbf{Pose Encoder:} Our pose encoder $Enc_{\textrm{pose}}$ takes two depth images and estimates the transformation in the latent space between two frames. Specifically, our pose encoder outputs 6D vectors representing the camera pose (rotation and translation) and we parameterize the pose encoder with a set of 2D convs. When constructing the \emph{NVM}, the pose encoder is used to estimate the transformation $T^{t}_{t}..T^{t}_{t-n+1}$ between previous frames and the present frame by comparing  previous visual observation $O^{v}_{t} .. O^{v}_{t-n+1}$ and present $O^{v}_{t}$.
\begin{equation*}
    T_{i}^{j} = Enc_{\textrm{pose}}(O^{v}_{i}, O^{v}_{j}) \in SE(3)
\end{equation*}

\noindent\textbf{Fusing 3D feature:} Given feature volumes from different frames and the estimated transformation from different frames to the present frame. We transform the feature volume $V_{t}..V_{t-n+1}$ with the corresponding transformation $T^{t}_{t} .. T^{t}_{t-n+1}$.  Particularly, for each position $p=(i,j,k)^{T}$ in 3D deep voxel, we transform it to $\hat{p}=(\hat{i},\hat{j},\hat{k})^{T}$ with:
\begin{equation*}
    \hat{p} = Rp + t
\end{equation*}
where $R$ is a 3x3 rotation matrix from $T^{t}_{t} .. T^{t}_{t-n+1}$ and $t$ is the translation vector from $T^{t}_{t} .. T^{t}_{t-n+1}$, and we note the transformation function as $f$.
\begin{equation*}    
\hat{V}_{i} =  f(V_{i}, T^{t}_{i}) \ \ \ \ i = t .. t - n + 1
\end{equation*}
Since directly performing transformation in latent space might cause misalignment due to the coarse of voxel representation, we use two additional 3D conv layers to refine the transformed feature volume.
We then aggregate transformed feature volumes to get Neural Volumetric Memory $\hat{V}^{m}_{t}$ representing the surrounding 3D structure in the present frame by computing the mean over feature volumes:
\begin{equation*}
\hat{V}^{m}_{t} = \frac{1}{n} \sum_{i=t-n+1}^{t} \hat{V}_{i}
\end{equation*}

\subsection{Learning NVM via Self-Supervision}

\begin{figure}[t]
\begin{center}
    \includegraphics[width=0.485\textwidth,clip]{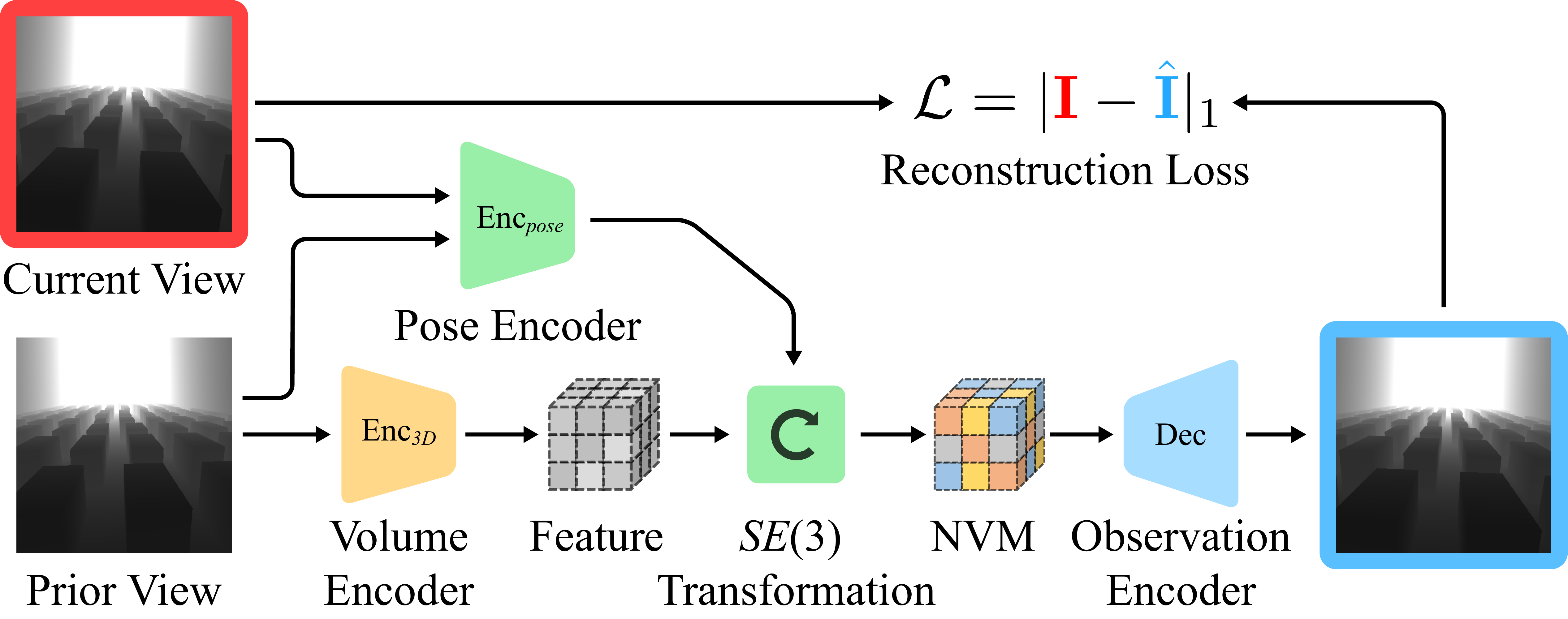}
\end{center}
    \vspace{-0.1in}
    \caption{\small{\textbf{Self-Supervised Learning:} we train a separate decoder to predict the visual observation in the different frames given one visual observation and the estimated transformation between two frames, similar to Lai et al.\cite{Lai21a}.}} 
    \label{fig:SSL}
\end{figure}

\noindent Although the behavior cloning objective is sufficient in producing a good policy, being equivariant to translation and rotation automatically offers a standalone, self-supervised learning objective just for the neural volumetric memory. Shown in Figure~\ref{fig:SSL}, one can assume that the surrounding 3D scene remains static between frames. Because the camera is forward-looking, we can canonicalized feature volume from earlier frames, and use it to predict a camera image taken at later steps. This self-consistency loss, computed as the pixel reconstruction error, encourages the memory to be $SE(3)$ equivariant.

For given visual observation history over a short span of time $O^{v}_{t} .. O^{v}_{t-n+1}$, we use the most dated visual observation  $O^{v}_{t-n+1}$ in the visual history and predict all frames. We extract a 3D feature volume from $O^{v}_{t-n+1}$: $V_{t-n+1} = Enc_{3D}(O^{v}_{t-n+1})$, estimate the relative camera pose $T^{t-n+1}_{t} .. T^{t-n+1}_{t-n+1}$ between $O^{v}_{t} .. O^{v}_{t-n+1}$ and the most dated $O^{v}_{t-n+1}$. We transform the extracted 3D feature volume $V_{t-n+1}$ into different frames with the corresponding transformations and then use a decoder $Dec$ to predict the visual observation $\hat{O}^{v}_{t..t-n+1}$ rendered in different frames. The Decoder $Dec$ takes a deep 3D feature volume as input and outputs a 2D image. The decoder first reshapes the deep 3D feature volume with shape $(C, D, H, W)$ into shape $(C*D, H, W)$ and applies a set of 2D conv layers to map the reshaped features to images. Given the predicted visual observation $\hat{O}^{v}_{t..t-n+1}$ and the original visual observation $O^{v}_{t} .. O^{v}_{t-n+1}$ in different frames, we compute the prediction loss with pixel-wise L1 loss. The 3D feature encoder and the pose encoder are both trained in this predictive self-supervised training task.
Specifically, we can formulate this 3D representation learning task as follows:
\begin{equation*}
\begin{aligned}
        V_{t-n+1} & = Enc_{3D}(O^{v}_{t-n+1}) \\
        \hat{O}^{v}_{i}  &= Dec(f(V_{t-n+1}, T_{i}^{t-n+1}))  \ \ i=t..t-n+1 \\
        L_{rec} & = \frac{1}{n} \sum_{i=t}^{t-n+1} | O^{v}_{i} - \hat{O}^{v}_{i} |
\end{aligned}
\end{equation*}
where $L_{rec}$ represents the L1 loss. Since the information contained in different visual observations is naturally different, perfect prediction in our task is not feasible. However, such a predictive learning task could still improve the 3D structural understanding of the surrounding and encourages the SE(3) equivariance over the learned 3D latent space, for this reconstruction loss encourages the disentanglement between the surrounding 3D structure and the camera pose. 

\subsection{Teacher Policy and Visuomotor Distillation}

\noindent Image observations introduce additional overhead
and slow training down as much as $40$ times\footnote{We are able to achieve 40k fps using just elevation map. Using $64\times 64$ images, it slows down to 1k fps.}.
For this reason, we bootstrap learning the visuomotor policy by first training a state-space policy that depends on the elevation map. 

\paragraph{Learning a state-only policy}
We use \textit{proximal policy optimization} (PPO~\cite{schulman2017proximal}) to train a single teacher policy $\hat{\pi}$ in all four environments. We parameterize the teacher policy as a 4-layer ReLU network with $[256, 128, 128, 64]$ latent neurons. To provide terrain information, we include two elevation maps, one larger and coarser than the other.
Overall, the observation space for the teacher policy include:
\begin{itemize}[leftmargin=*]
    \setlength\itemsep{0em}
    \item \textbf{proprioceptive state - $\mathbb R^{63}$} A $63$-dimension vector consists of the gravity vector in the body frame $\textbf{g}_{t}$, joint rotations $\textbf{q}_{t}$, joint velocity $\dot{\textbf{q}}_{t}$, and previous actions $\textbf{a}_{t-1}$.
    \item \textbf{privileged information - $\mathbb R^{14}$} A $14$-dimension vector including linear velocity $\textbf{v}_{t}$ and angular velocity $\boldsymbol{\omega}_{t}$ of the robot, and environment parameters $\textbf{e}_{t}$.
    \item \textbf{a dense elevation map $m_{dense}$ - $\mathbb R^{21\times 26}$} with $5\cm$ spacing between the grid points, which is only slightly bigger than the robot.
    \item \textbf{a sparse elevation map $m_{sparse}$ - $\mathbb R^{10\times 19}$} with $10\cm$ spacing. This covers a much larger area.
\end{itemize}
We provide a more detailed description of the observation in the appendix.

\paragraph{Reward function}: We use the same reward function for teacher's  training in all environments. The reward contains the following terms: 1) Reward for moving forward at a target velocity (in our experiment: 0.4m/s). 2) an energy penalty to encourage the robot to use as little torque as possible.  3) Height tracking reward encouraging the robot to walk while maintaining a specific relative height to the ground (in our experiment: 0.265m). 4) AMP reward ~\cite{2021-TOG-AMP} encouraging the agent to produce natural gaits.
\begin{equation*}
\begin{aligned}
 R =&\; \alpha_\text{forward} * R_\text{forward} + \alpha_\text{energy} * R_\text{energy} \\
   &+\; \alpha_\text{height} * R_\text{height} + \alpha_\text{amp} * R_\text{amp}
\end{aligned}
\end{equation*}
We include the details of each reward term and the corresponding weights in the appendix.

\paragraph{Visuomotor distillation:}  
Given the privileged teacher policy from stage one, we then distill the privileged teacher policy $\hat{\pi}$ into the visual policy $\pi$ without privileged information. Our visual policy utilizes proprioceptive state and visual observation for decision-making. The proprioceptive state $O_{prop}$ is the same as we used to train privileged elevation-based policy in stage one while the visual observation contains a sequence of depth images ($O^{v}_{t} .. O^{v}_{t-n+1}$, n=5 in our experiment) with a shape 64 × 64 from a depth camera mounted on the head of the robot. The observation space comparison between the privileged teacher policy and the visual policy is provided in Table~\ref{table:observation_comparison}, and the specific position of the camera is provided in Figure~\ref{fig:sim2real_pipeline}.

\begin{table}[t]
\centering
\caption{\textbf{Comparison between observation spaces}. State-only policy (w/ elevation-map) vs. vision policy}
\vspace{-0.10in}

\begin{small}
\begin{tabular}{c|c|c|c|c}
\toprule
 & \rotatebox{-0}{\makecell{Joint\\states} }
 & \rotatebox{-0}{\makecell{Velocity} }
 & \rotatebox{-0}{\makecell{Elevation\\map} }
 & \rotatebox{-0}{\makecell{Depth\\map} } \\
\midrule
 Teacher policy  
    & {\color{es-blue}\Checkmark}  
    & {\color{es-blue}\Checkmark} 
    & {\color{es-blue}\Checkmark} 
    & {\color{lightgray}\XSolidBrush} \\
Visual policy 
    & {\color{es-blue}\Checkmark}  
    & {\color{lightgray}\XSolidBrush} 
    & {\color{lightgray}\XSolidBrush} 
    & {\color{es-blue}\Checkmark} \\ 
\bottomrule
\end{tabular}
\end{small}
\vspace{-0.05in}
\label{table:observation_comparison}
\end{table}

As shown in Figure~\ref{fig:NVM}, our visual policy first processes the sequential visual observation with our proposed NVM module. After obtaining NVM with shape $(C, D, H, W)$, our visual policy flattens the NVM into a 2D feature map with shape $(C \times D, H, W)$ and utilizes a set of 2D conv layers and a single linear layer to get a 1D visual feature. Our visual policy encodes the proprioceptive state with a neural network and gets the proprioceptive feature. The proprioceptive feature and visual feature are then concatenated and fed into a final network to get the actual action for the robot.

The visual policy is trained with behavior cloning by minimizing the L1 distance between actions from teacher policy $\hat{\pi}$ and visual policy $\pi$,
\begin{equation*}
    L_{BC} = |\hat{\pi}(p) - \pi(o)|
\end{equation*}
where $\mathbf{p}$ is the privileged observation (including elevation maps), and $\mathbf{o}$ is the observation (including visual observations). Combined with our proposed self-supervised task, the overall training loss for our NVM is given by:
\begin{equation*}
    L = \lambda_{BC} L_{BC} + \lambda_{rec} L_{rec}
\end{equation*}
In our experiment, we use $\lambda_{BC}=1$ and $\lambda_{rec}=0.01$.

\section{Experiments}
\label{sec:experiment}

\begin{table*}[t]
\caption{\small{{\bf Evaluation of All Policies:} We evaluate all methods in all four environments and show their Traversing Rate and Success Rate. The privileged elevation-map-based policy is noted as Teacher, and its performance is regarded as the upper bound for the rest methods. }} 

\label{table:student_evaluation}
\centering
\resizebox{\textwidth}{!}{
\begin{tabular}{lcccccccc}
\toprule

& \multicolumn{4}{c}{Traversing Rate (\%) $\uparrow$ }  & \multicolumn{4}{c}{Success Rate (\%) $\uparrow$ }\\
\midrule
Scenarios& Stages& Stairs& Stones & Obstacles& Stages& Stairs& Stones & Obstacles\\\midrule
Teacher& $\mathbf{97.9 \scriptstyle{\pm 2.0 }}$ & $\mathbf{94.5 \scriptstyle{\pm 1.8 }}$ & $\mathbf{89.6 \scriptstyle{\pm 3.1 }}$ & $\mathbf{92.8 \scriptstyle{\pm 1.8 }}$ & $\mathbf{95.6 \scriptstyle{\pm 2.3 }}$ & $\mathbf{86.6 \scriptstyle{\pm 2.7 }}$ & $\mathbf{70.6 \scriptstyle{\pm 5.2 }}$ & $\mathbf{87.8 \scriptstyle{\pm 2.7 }}$ \\
NaiveCNN& $76.4 \scriptstyle{\pm 26.9 }$ & $81.5 \scriptstyle{\pm 28.6 }$ & $29.4 \scriptstyle{\pm 10.6 }$ & $63.3 \scriptstyle{\pm 22.5 }$ & $53.9 \scriptstyle{\pm 19.7 }$ & $71.3 \scriptstyle{\pm 25.4 }$ & $0.5 \scriptstyle{\pm 0.8 }$ & $45.2 \scriptstyle{\pm 16.4 }$ \\
NaiveCNN-MS& $72.1 \scriptstyle{\pm 25.4 }$ & $84.5 \scriptstyle{\pm 29.7 }$ & $26.4 \scriptstyle{\pm 9.3 }$ & $63.7 \scriptstyle{\pm 22.7 }$ & $48.9 \scriptstyle{\pm 18.9 }$ & $73.1 \scriptstyle{\pm 26.2 }$ & $0.0 \scriptstyle{\pm 0.0 }$ & $48.1 \scriptstyle{\pm 18.3 }$ \\
NaiveCNN-RNN& $68.3 \scriptstyle{\pm 24.1 }$ & $82.4 \scriptstyle{\pm 29.7 }$ & $26.3 \scriptstyle{\pm 9.3 }$ & $67.9 \scriptstyle{\pm 23.9 }$ & $48.9 \scriptstyle{\pm 18.0 }$ & $79.1 \scriptstyle{\pm 28.9 }$ & $2.2 \scriptstyle{\pm 1.5 }$ & $49.8 \scriptstyle{\pm 17.9 }$ \\
LocoTransformer& $74.0 \scriptstyle{\pm 26.5 }$ & $85.5 \scriptstyle{\pm 29.9 }$ & $25.2 \scriptstyle{\pm 8.9 }$ & $65.2 \scriptstyle{\pm 22.9 }$ & $53.3 \scriptstyle{\pm 19.3 }$ & $75.8 \scriptstyle{\pm 26.8 }$ & $0.9 \scriptstyle{\pm 1.1 }$ & $46.7 \scriptstyle{\pm 16.9 }$ \\
NVM& $\mathbf{79.9 \scriptstyle{\pm 28.1 }}$ & $\mathbf{85.5 \scriptstyle{\pm 30.0 }}$ & $\mathbf{47.4 \scriptstyle{\pm 16.7 }}$ & $\mathbf{74.1 \scriptstyle{\pm 26.0 }}$ & $\mathbf{65.0 \scriptstyle{\pm 23.2 }}$ & $\mathbf{79.8 \scriptstyle{\pm 28.2 }}$ & $\mathbf{14.4 \scriptstyle{\pm 6.0 }}$ & $\mathbf{61.6 \scriptstyle{\pm 21.9 }}$ \\

\bottomrule

\end{tabular}
}
\end{table*}

\subsection{Simulation Environments}
\label{sec:simulation}

\noindent We evaluate NVM on four challenging terrains shown in Figure~\ref{fig:sim2real_pipeline}. Our intention is to evaluate the global and local perception ability of agents, as well as the robustness of the learned locomotion skills. Our environments consist of \textbf{Stages} including multiple separate stages with gaps of different sizes between stages; \textbf{Obstacles} including multiple tall cuboids for the agent to avoid; \textbf{Stairs} including stairs with different heights and widths; \textbf{Stones} including numerous small stones of different shapes for the agent to step over. \textbf{Stages}, \textbf{Stairs}, and \textbf{Stones} environments evaluate the agent's ability to reason about the nearby surrounding and perform precise foot placement. \textbf{Obstacle} environment requires the agent to be able to perform long-term planning and avoid obstacles beforehand.

\subsection{Baselines}

\noindent We compare our method against baselines that do not contain SE(3) equivariance priors: 

\begin{itemize}[leftmargin=*]
    \setlength\itemsep{0em}
    \item\textbf{NaiveCNN} which learns visual features with a 2D CNN encoder followed by a linear projection. Visual features and proprioceptive features are concatenated. The action is then computed from the concatenated feature.

    \item\textbf{NaiveCNN-RNN} which shares a similar structure with \emph{NaiveCNN}, but before computing action from the concatenated feature, \emph{NaiveCNN-RNN} baseline utilizes a 2-layer GRU to provide memory mechanism for the agent

    \item\textbf{LocoTransformer} which utilizes a shared transformer model to perform cross-modal reasoning between visual tokens from 2D visual feature map and proprioceptive features, instead of directly concatenating visual feature and proprioceptive feature.

    \item\textbf{Multi-Step NaiveCNN (NaiveCNN-MS)} which use a 2D CNN to extract visual feature from each depth image and fuse them together to get the overall visual feature, instead of stacking multiple depth images and processing the stacked visual observation with a 2D CNN.
 
\end{itemize}

For \emph{NaiveCNN} and \emph{LocoTransformer}, the visual observations in the near past are stacked in the channel dimension and serve as input for 2D encoder. All these baselines and our method share the same 2D CNN encoder structure for a fair comparison.

\subsection{Simulated Results}

\begin{table*}[t]
\caption{\textbf{Ablation study}. We scale the grid resolution of the neural volumetric memory and the horizon of the past kept in memory. By default, our NVM uses use $D=12, H=W=6, n=5$.}

\label{table:ablation_study}
\centering

\resizebox{\textwidth}{!}{
\begin{tabular}{lcccccccc}
\toprule

& \multicolumn{4}{c}{Traversing Rate (\%) $\uparrow$ }  & \multicolumn{4}{c}{Success Rate (\%) $\uparrow$ }\\
\midrule
Scenarios& Stages& Stairs& Stones& Obstacles& Stages& Stairs& Stones& Obstacles\\\midrule
NVM& $\mathbf{79.9 \scriptstyle{\pm 28.1 }}$ & $85.5 \scriptstyle{\pm 30.0 }$ & $47.4 \scriptstyle{\pm 16.7 }$ & $\mathbf{74.1 \scriptstyle{\pm 26.0 }}$ & $\mathbf{65.0 \scriptstyle{\pm 23.2}}$ & $79.8 \scriptstyle{\pm 28.2 }$ & $14.4 \scriptstyle{\pm 6.0 }$ & $\mathbf{61.6 \scriptstyle{\pm 21.9 }}$ \\

\midrule
NVM(D=3)& $65.4 \scriptstyle{\pm 23.7 }$ & $68.0 \scriptstyle{\pm 25.3 }$ & $38.6 \scriptstyle{\pm 14.5 }$ & $46.5 \scriptstyle{\pm 17.0 }$ & $52.8 \scriptstyle{\pm 19.4 }$ & $52.2 \scriptstyle{\pm 19.7 }$ & $5.9 \scriptstyle{\pm 4.1 }$ & $36.1 \scriptstyle{\pm 13.5 }$ \\
NVM(D=6)& $79.3 \scriptstyle{\pm 27.8 }$ & $83.1 \scriptstyle{\pm 29.2 }$ & $\mathbf{49.8 \scriptstyle{\pm 17.6 }}$ & $66.2 \scriptstyle{\pm 23.6 }$ & $64.2 \scriptstyle{\pm 23.0 }$ & $74.7 \scriptstyle{\pm 26.4 }$ & $\mathbf{14.7 \scriptstyle{\pm 6.2 }}$ & $50.2 \scriptstyle{\pm 19.3 }$ \\

\midrule
NVM(D=4x4)& $72.8 \scriptstyle{\pm 26.5 }$ & $70.1 \scriptstyle{\pm 25.9 }$ & $39.1 \scriptstyle{\pm 14.4 }$ & $66.4 \scriptstyle{\pm 24.0 }$ & $54.8 \scriptstyle{\pm 21.4 }$ & $62.8 \scriptstyle{\pm 23.7 }$ & $11.3 \scriptstyle{\pm 6.0 }$ & $48.6 \scriptstyle{\pm 18.4 }$ \\
NVM(D=8x8)& $78.9 \scriptstyle{\pm 27.8 }$ & $73.4 \scriptstyle{\pm 26.3 }$ & $41.8 \scriptstyle{\pm 15.0 }$ & $64.5 \scriptstyle{\pm 22.9 }$ & $61.4 \scriptstyle{\pm 22.3 }$ & $64.8 \scriptstyle{\pm 23.9 }$ & $12.2 \scriptstyle{\pm 4.8 }$ & $50.0 \scriptstyle{\pm 18.3 }$ \\

\midrule
NVM(n=3)& $70.1 \scriptstyle{\pm 24.8 }$ & $76.6 \scriptstyle{\pm 26.9 }$ & $37.6 \scriptstyle{\pm 13.3 }$ & $64.3 \scriptstyle{\pm 23.0 }$ & $43.9 \scriptstyle{\pm 16.8 }$ & $67.0 \scriptstyle{\pm 23.9 }$ & $0.6 \scriptstyle{\pm 0.8 }$ & $48.9 \scriptstyle{\pm 18.1 }$ \\
NVM(n=9)& $78.4 \scriptstyle{\pm 27.6 }$ & $\mathbf{85.9 \scriptstyle{\pm 30.1}}$ & $39.8 \scriptstyle{\pm 14.1 }$ & $69.8 \scriptstyle{\pm 24.6 }$ & $61.6 \scriptstyle{\pm 22.0 }$ & $\mathbf{80.0 \scriptstyle{\pm 28.5 }}$ & $5.8 \scriptstyle{\pm 3.2 }$ & $56.1 \scriptstyle{\pm 19.9 }$ \\

\bottomrule
\end{tabular}
}
\end{table*}

 \noindent\textbf{Evaluation metrics:} We evaluate the final policies of all methods for 640 episodes in each simulated environment and compare the performance. We evaluate different methods by two metrics: \textbf{traversing rate(\%)} and \textbf{success rate(\%)}. The traversing rate is defined by the distance the robot moves dividing the reachable distance (the distance between the end of the environment and starting position), and the Success Rate is defined by the ratio of the robot reaching the end of the environment.

We evaluated all methods including the privileged teacher policy (noted as Teacher) across all environments and the results are provided in Table~\ref{table:student_evaluation}. Since in stage two of our pipeline, visual policies are trained by imitation learning (our method with self-supervised learning as well), the performance of the privileged teacher policy can be regarded as the upper bound for the performance of visual policies. We can see that our NVM outperforms all the baselines in all the environments, especially the Stones environment which requires the agent to estimate the 3D structure of the surroundings from visual observations and provide accurate foot placement. Surprisingly, it turns out our method traversed a longer distance in the obstacle environment, where the understanding of 3D structure is not required since it's possible to avoid the obstacles with only the latest visual observation. In this case, we conjecture that the 3D understanding of the environment helps the agent to keep closer to the trajectory of the teacher policy would take since the teacher could exploit the 3D structure of the surroundings to find the optimal path.

We find that though \emph{NaiveCNN-MS} introduces the mechanism of processing each frame of visual observation separately, there is no improvement compared with \emph{NaiveCNN}. This comparison indicates that the improvement brought by our NVM is actually from explicitly modeling the 3D transformation instead of introducing more computation. Compared with \emph{NaiveCNN}, the memory-enabled \emph{NaiveCNN-RNN} also shows no improvement, which indicates that the memory mechanism provided by RNN is not better than, if not worse, naively stacking the visual observations in the near past.

\subsection{Real World Results}

\noindent To validate the performance of our method beyond simulation, we constructed multiple real-world experiment scenes as shown on the right of  Figure~\ref{fig:sim2real_pipeline}. We evaluated our method in three scenes: \textbf{Stages:} we build $4$ discrete stages with \(15 ~\cm\) / \(20 ~\cm\) / \( 30 ~\cm\) between stages. \textbf{Stairs:} We build a stair consisting of $3$ ladders with heights of \(3 ~\cm\) / \(6 ~\cm\) / \(10 ~\cm\). \textbf{Obstacles:} We place multiple cuboid obstacles in the narrow hallway for the robot to avoid. We deploy the policies in real-world scenes and record the distance robot moved for each episode. We terminate the episode after a certain amount of time or until the robot reaches the end of the environment or the robot falls. We repeat the experiment for each method and environment five times. Real-world trajectories are also provided in Figure~\ref{fig:teaser}. Since \emph{NaiveCNN}, \emph{NaiveCNN-RNN}, \emph{LocoTransformer} show similar performance in simulation and \emph{LocoTransformer} is known to be more robust in unseen scenarios and objects in the real world, we compared our NVM with \emph{LocoTransformer} in these three scenarios. The quantitative results are shown in Table~\ref{table:real_world_results}. We find that our NVM moved significantly further in \emph{Stages} and \emph{Stairs} while slightly further in \emph{Obstacles}. This is because \emph{Stages} and \emph{Stairs} environment require the robot to understand the surrounding 3D structure to place the foot while \emph{Obstacles} only requires the ability to perform high-level planning with visual observation. We observe that in \emph{Stages} and \emph{Stairs} environment, agents trained with \emph{LocoTransformer} can place the front leg across the stage or over the ladder, but not the rear legs when the gap between stages is large or the ladder is tall, which demonstrates that lack of  understanding of the surrounding 3D structure. We also found that even though \emph{LocoTransformer} agent could move a reasonable distance in simulation in \emph{Stairs} and \emph{Stages} environment, it fails in the real world. The real-world results indicate that the understanding of the surrounding 3D structures is essential for legged robots to traverse complex terrains in the real world and our NVM module provides a high-quality volumetric memory of the surrounding 3D structure.

\begin{table}[t]
\caption{Evaluating the visual policy in the real world}
\label{table:real_world_results}
\centering
\begin{tabular}{lcccc}
\toprule
\multicolumn{4}{c}{Distance Moved (\m) $\uparrow$ }\\
\midrule
Scenarios& Stages& Stairs& Obstacles\\\midrule
NVM& $\mathbf{5.4 \scriptstyle{\pm 0.3 }}$ & $\mathbf{4.0 \scriptstyle{\pm 0.0 }}$ & $\mathbf{7.3 \scriptstyle{\pm 0.1 }}$ \\
LocoTransformer& $3.4 \scriptstyle{\pm 0.1 }$ & $2.1 \scriptstyle{\pm 0.8 }$ & $6.7 \scriptstyle{\pm 0.4 }$ \\

\bottomrule
\end{tabular}

\vspace{-0.1in}
\end{table}

\begin{figure*}[t]
    \begin{subfigure}[a]{0.49\textwidth}
        \includegraphics[width=\textwidth,clip]{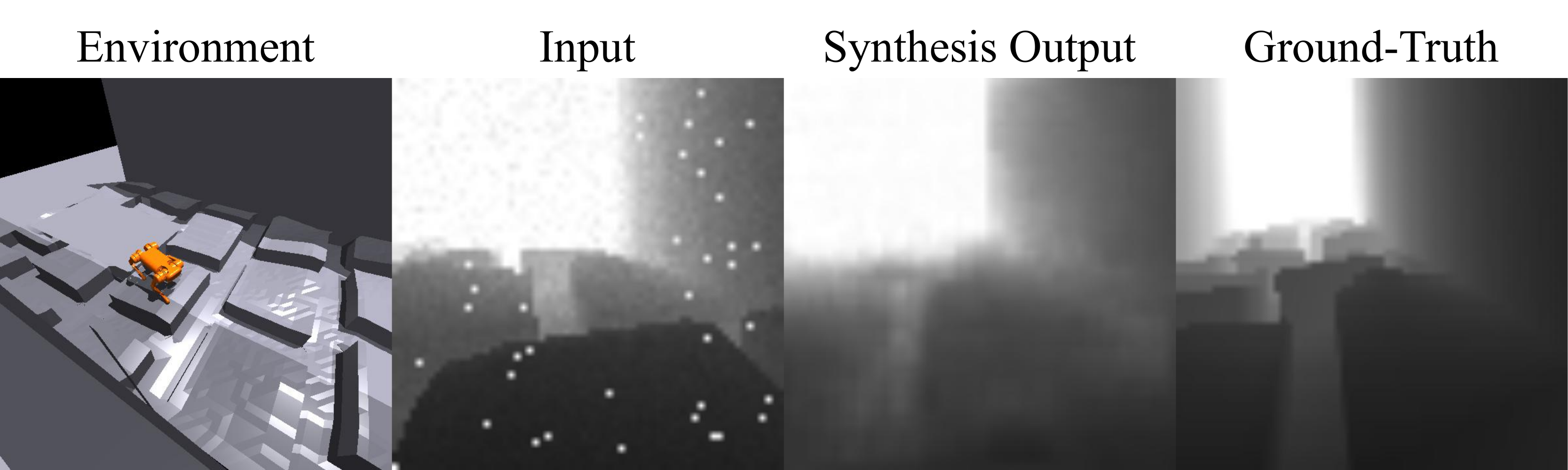}
        \caption{Stages}
        \label{fig:ssl_visualization_stages}
    \end{subfigure}\hfill
    \begin{subfigure}[a]{0.49\textwidth}
        \includegraphics[width=\textwidth,clip]{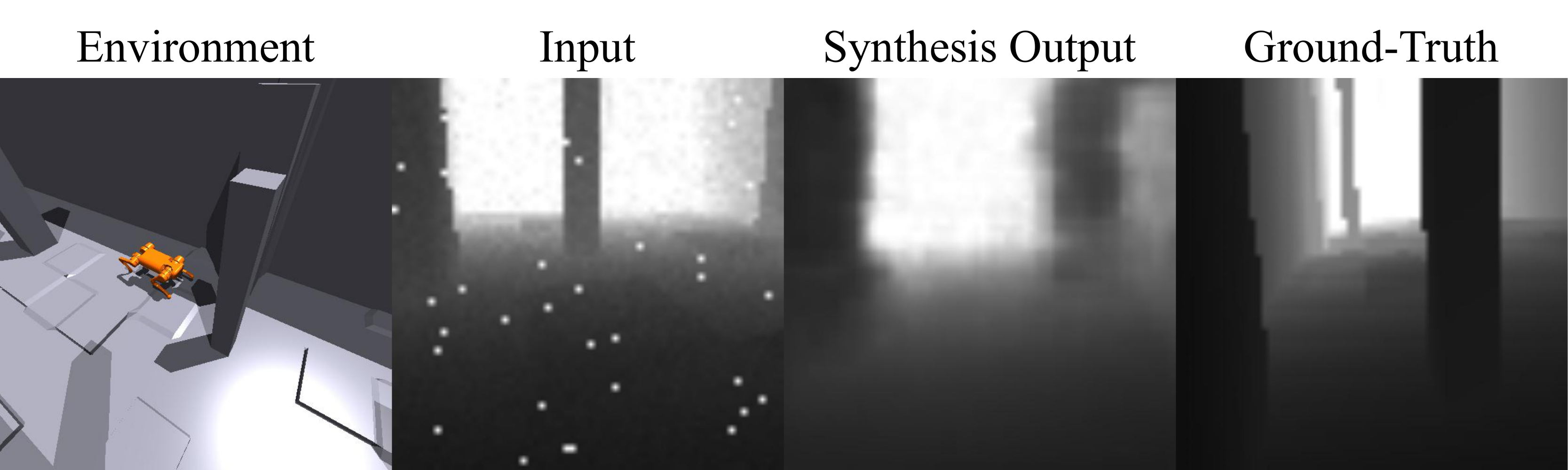}
        \caption{Obstacles}
        \label{fig:ssl_visualization_osbtacles}
    \end{subfigure}\hfill
    
    \begin{subfigure}[a]{0.49\textwidth}
        \includegraphics[width=\textwidth,clip]{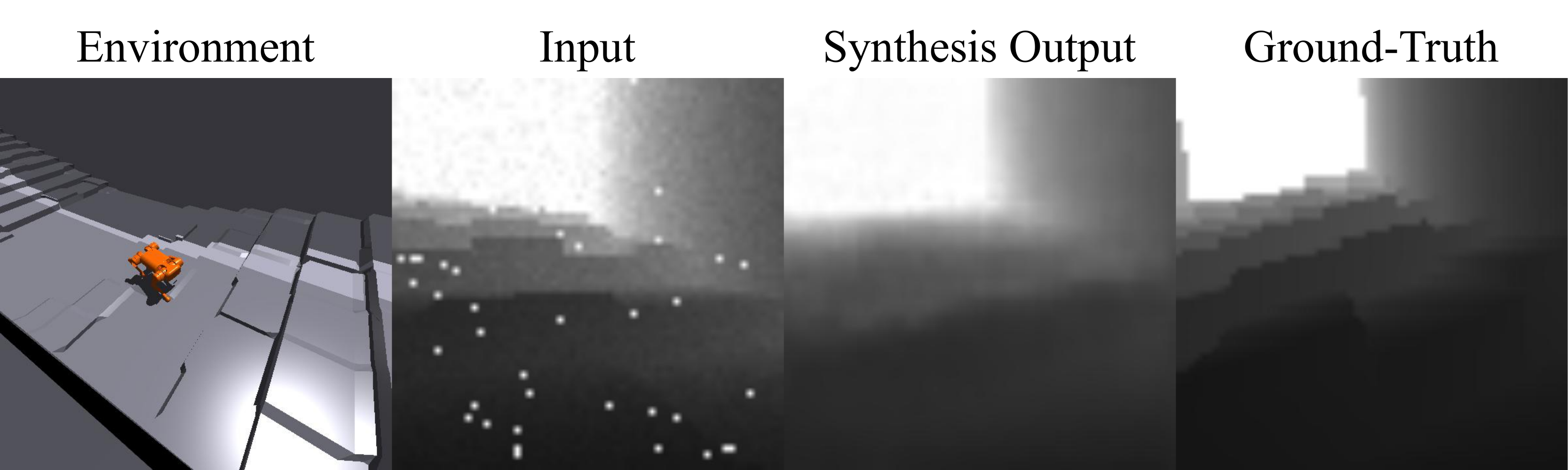}
        \caption{Stairs}
        \label{fig:ssl_visualization_stairs}
    \end{subfigure}\hfill
    \begin{subfigure}[a]{0.49\textwidth}
        \includegraphics[width=\textwidth,clip]{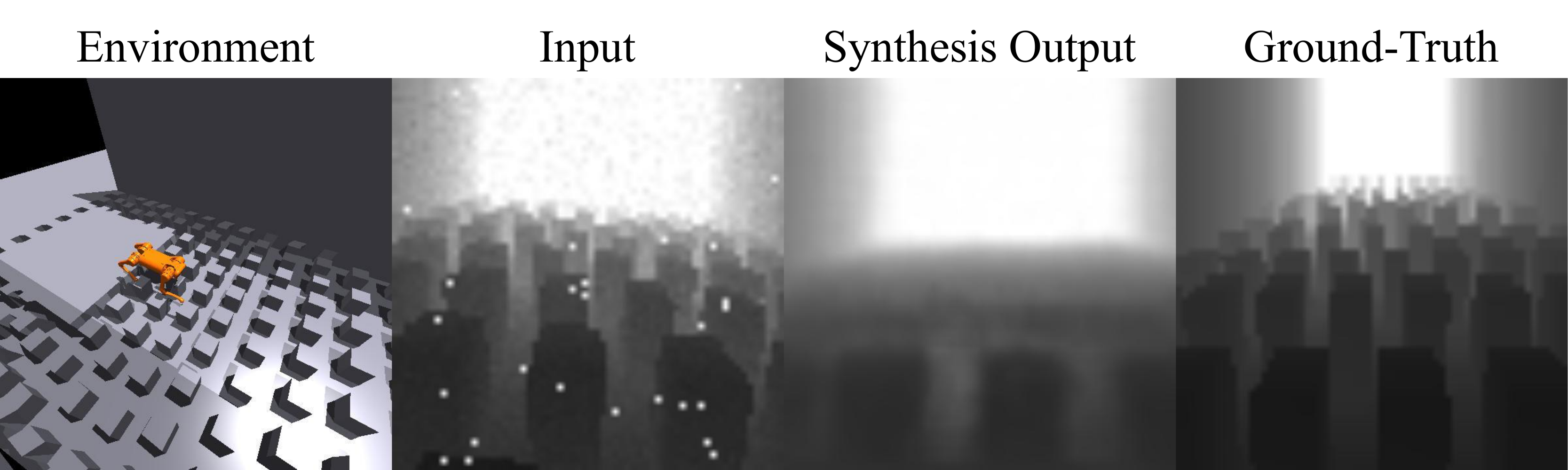}
        \caption{Stones}
        \label{fig:ssl_visualization_stepping_stones}
    \end{subfigure}\hfill
    
    \vspace{-0.05in}
    \caption{
        \textbf{Visual reconstruction from learned decoder}. We visualize the synthesized visual observation in our self-supervised task. For every tuple, the first image shows the robot moving in the environment, the second image is the input visual observation $O^{v}_{t-n+1}$, the third image is the synthesized visual observation using 3D feature volume $V_{t-n+1}$ extracted from $O^{v}_{t-n+1}$ and the estimated relative camera between $O^{v}_{t-n+1}$ and robot's ego-centric frame, and the fourth image is the actual visual observation captured by the robot (the ground-truth in our self-supervised task). For the input visual observation, we apply extensive data augmentation to the image to improve the robustness of our model.
    }
    \label{fig:ssl_visualization}
    \vspace{-0.1in}
\end{figure*}

\subsection{Visualization of SSL task}
\noindent Aiming to gather insight into our learned 3D features volume, we visualized the synthesized visual observation $\hat{O^{v}_{t}}$ in our self-supervised learning task along with the present visual observation $O^{v}_{t}$ (the ground truth in our SSL task), and most dated visual observation $O^{v}_{t-n+1}$  (the input in our SSL task) in Figure~\ref{fig:ssl_visualization}. Even though we apply extensive data augmentation over the raw visual observation (the second image for every case), the synthesized visual observations are clearly indicating the robustness of our model. In the synthesized visual observation for \emph{Stair} and \emph{Obstacle} environment, we can see the overall structure of the environment is well-preserved. In Figure~\ref{fig:ssl_visualization_osbtacles}, The obstacles in the previous frame are transformed to the position in the current frame without much distortion. In Figure~\ref{fig:ssl_visualization_stairs}, the boundary of the nearest ladder is clear, and the second nearest ladder is also observable in the present frame. We have to note that the reason the decoder can not recover the details away from the robot is in the input for SSL the large values are clipped. For more complex \emph{Stages} and \emph{Stones} environments, we found that even though the terrain details are not perfectly reconstructed, the essential regions for decision-making are reconstructed which means the learned 3D voxel features capture the most important parts in the visual observations. In Figure~\ref{fig:ssl_visualization_stages}, we can see the gaps between stages nearby are well reconstructed and in Figure~\ref{fig:ssl_visualization_stepping_stones}, we can see the reconstructed stone in front of the robot.

\subsection{Ablation Study}
\noindent we perform multiple sets of ablation studies, and the results are shown in Table~\ref{table:ablation_study}, with the purpose of studying the influence of different components in our method.

\noindent\textbf{The Resolution of NVM:} We provide our NVM module voxel features with different resolutions by slightly changing the structure of the 2D encoder, which resulting 2D feature maps in different shapes and use different depth parameter $D$. From Table~\ref{table:ablation_study}, we find that the larger depth value $D$ generally provides better performance, since larger $D$ converts the 2D feature map into more fine-grained voxel feature which results in fine-grained neural volumetric memory, and it's supposed to be easier for the agent to reason about the foot placement over complex terrain with more fine-grained neural volumetric memory. For $H$ and $W$, we found that $6x6$ which provides enough detail horizontally without introducing much noise vertically works best.

\noindent\textbf{The Length of Visual History of NVM:} We provide our NVM model with different lengths of visual history by changing the number of stored visual observation frames. As shown in Table~\ref{table:ablation_study}, the performance of \emph{NVM} improves when the length of visual history increases to 5 from 3, we think this is because the longer visual history provides longer 3D memory of the surrounding for the agent to safely placing the rear feet. But when the length of visual history increases to 9 from 5, the performance drops, we hypothesize that it is because when the visual history becomes too long, it's hard for the pose encoder to estimate the accurate transformation between the oldest frame and the current frame, and the surplus visual information might be redundant for decision-making over terrains.

\section{Conclusion}
\label{sec:discussion}

\noindent We propose a short-term memory mechanism, the Neural Volumetric Memory (NVM), to enable legged robots to traverse complex terrains with ease. NVM aggregates knowledge from sequential visual observation in the near past by explicitly modeling the translation and rotation between previous frames and the present frame. The explicit modeling of the transformation and self-supervised task also encourage the $SE(3)$ equivariance in the memory space. Our neural volumetric memory module enables the legged robot to traverse complex terrain in simulation and the real world and achieve better sim-to-real generalization results. This shows our neural volumetric memory provides reliable surrounding 3D structure information for the robot to perform decision-making and new possibility of utilizing visual observation for legged-robot.

{\noindent \small \textbf{Acknowledgements.}  This project was supported, in part, by NSF CCF-2112665 (TILOS), NSF CAREER Award IIS-2240014, NSF 1730158 CI-New: Cognitive Hardware and Software Ecosystem Community Infrastructure (CHASE-CI), NSF ACI-1541349 CC*DNI Pacific Research Platform, Amazon Research Award and gifts from Qualcomm. 
Ge Yang is supported by the
National Science Foundation Institute for Artificial Intelligence
and Fundamental Interactions (IAIFI, https://iaifi.org/) under
the Cooperative Agreement PHY-2019786. We thank Yuxiang Yang for providing help regarding hardware setup.}

{\small
\bibliographystyle{ieee_fullname}
\bibliography{egbib}
}

\newpage
\begin{center}
\large \bf Appendix \par
\end{center}

\setcounter{section}{0}
\section{Detailed Observation Space}

\label{sec:appendix_obs}
\noindent In this section, we provide more details about different components of observation space for both the privileged teacher policy and visual policies. 
The Unitree A1 robot we used has 12 joints, corresponding to 12 Degrees of Freedom (DoF), and we use positional control for the 12 DOF. Specifically, the proprioceptive input contains: 

\begin{itemize}
    \item \textbf{Joint Rotations - $\mathbb R^{12 \times 2}$} contains joint rotations for all joints ($12$D) for the past two control step. 
    \item \textbf{Joint Velocity - $\mathbb R^{12}$} contains joint velocities for all joints ($12$D).
    \item \textbf{Previous Action - $\mathbb R^{12\times 2}$} contains positional command for all joints ($12$D) for the past two control step.
    \item \textbf{Projected Gravity - $\mathbb R^{3}$} contains the projected gravity in the robot frame, representing the orientation of the robot.
\end{itemize}

The privileged information for privileged teacher policy training contains:

\begin{itemize}
    \item \textbf{Linear Velocity - $\mathbb R^{3}$} contains the linear velocity of the robot in the world frame. 
    \item \textbf{Angular Velocity - $\mathbb R^{3}$} contains the linear velocity of the robot in the world frame.
    \item \textbf{Environment Parameters - $\mathbb R^{8}$} contains randomized environment parameters.
\end{itemize}

For visual observation, we provide $N=5$ frames of the depth image ($64 \times 64$) to construct the perception history. To simulate the noisy visual observation in the real world, for each time step, we randomly sample $1\%$ pixels in $(64, 64)$ depth image and set the reading for these pixels to be the maximum reading. 

\section{Reward for Privileged Teacher Training}
\label{sec:appendix_rew}

\noindent For the training of privileged teacher policy in all environments we use the same reward function as follows: 
\begin{equation*}
\begin{aligned}
 R =&\; \alpha_\text{forward} * R_\text{forward} + \alpha_\text{energy} * R_\text{energy} \\
   &+\; \alpha_\text{height} * R_\text{height} + \alpha_\text{amp} * R_\text{amp}
\end{aligned}
\end{equation*}
In our experiment, we use $\alpha_{\text{forward}} = 1, \alpha_{\text{energe}} =  -0.005, \alpha_{\text{height}} = -2.0, \alpha_{\text{amp}} = 1.0$

we provide specific formulations of different reward terms in our reward function

\begin{equation*}
\begin{aligned}
 R_{\text{forward}} = 1 + | v_{\text{robot}} - v_{\text{target}} | / v_{\text{target}}
\end{aligned}
\end{equation*}
where $v_{\text{robot}}$ is the current robot speed along the forward direction, and the $v_{\text{target}} = 0.4$ is the target moving forward speed.

\begin{equation*}
\begin{aligned}
 R_\text{energy} = \sum_{i} |\tau_{i} \times \dot{q}_{i}|
\end{aligned}
\end{equation*}
where $\tau_{i}$ is the the motor torques applied to the $i$th joint, and the $\dot{q}_i$ is the joint velocity for the $i$th joint.

\begin{equation*}
\begin{aligned}
 R_\text{height} = || h_{\text{robot}} - h_{\text{target}}||
\end{aligned}
\end{equation*}
where $h_{\text{robot}}$ is the current relative height of the robot with respect to the terrain, and the $h_{\text{target}} = 0.265$ is the target height to track.

For AMP reward, we follow the setting in Peng et al ~\cite{2021-TOG-AMP}, where a gait discriminator is trained to distinguish the gait in the reference motions and the gait produced by the RL policy. The score (between $0$ and $1$) from the discriminator is used as the AMP reward. The gait closer to the reference motions is assigned a higher reward.

\end{document}